\documentclass[letterpaper, 10 pt, conference]{ieeeconf}
\IEEEoverridecommandlockouts 
\UseRawInputEncoding

\usepackage{amsmath,amsfonts,amssymb}
\usepackage{algorithmic}
\usepackage{array}
\usepackage[caption=false,font=normalsize,labelfont=sf,textfont=sf]{subfig}
\usepackage{textcomp}
\usepackage{stfloats}
\usepackage{matlab-prettifier}
\usepackage{url}
\usepackage{verbatim}
\usepackage{float}
\usepackage{graphicx}
\usepackage{balance}
\usepackage{tabularx}
\usepackage{bm}
\usepackage{cite}
\usepackage{caption}
\usepackage{mathtools}
\usepackage{footnote}
\usepackage{diagbox}
\usepackage{bm}
\usepackage{subfig}
\usepackage{booktabs}
\usepackage{xcolor}
\let\labelindent\relax      
\usepackage{enumitem}
\usepackage[hidelinks]{hyperref}
\usepackage{xcolor,soul,lipsum}

\newcolumntype{Y}{>{\centering\arraybackslash}X}

\begin{document}
\title{Exp[licit]\\
\vspace{2mm}
\large A Robot modeling Software based on Exponential Maps}
\author{Johannes Lachner, Moses C. Nah, Stefano Stramigioli, Neville Hogan}
\author{Johannes Lachner$^{*}$, Moses C. Nah$^{*}$, Stefano Stramigioli
\IEEEmembership{Fellow, IEEE}, Neville Hogan \IEEEmembership{Member, IEEE}
\thanks{$^{*}$J. Lachner and M. C. Nah contributed equally}
\thanks{J. Lachner, M. C. Nah, and N. Hogan are with the Department of Mechanical Engineering, Massachusetts Institute of Technology,
Cambridge, MA 02139 USA}
\thanks{N. Hogan is also with the Department
of Brain and Cognitive Sciences, Massachusetts Institute of Technology,
Cambridge, MA 02139 USA}
\thanks{S. Stramigioli is with the Faculty of Electrical Engineering, Mathematics and Computer Science, University of Twente, 7522 Enschede,
The Netherlands}
\thanks{This work has been submitted to the IEEE for possible publication. Copyright may be transferred without notice, after which this version may no longer be accessible.}
}

\markboth{IEEE Robotics $\&$ Automation Magazine, August~2022}
{J. Lachner, M. C. Nah, \MakeLowercase{\textit{(et al.)}:
Robot Modeling with Exponential Maps}}

\maketitle

\begin{abstract}
Deriving a robot's equation of motion typically requires placing multiple coordinate frames, commonly using the Denavit-Hartenberg convention  to express the kinematic and dynamic relationships between segments. This paper presents an alternative using the differential geometric method of Exponential Maps, which reduces the number of coordinate frame choices to two. The traditional and differential geometric methods are compared, and the conceptual and practical differences are detailed. The open-source software, Exp[licit]\textsuperscript{TM},  based on the differential geometric method, is introduced. It is intended for use  by researchers and engineers with basic knowledge of geometry and robotics. Code snippets and an example application are provided to demonstrate the benefits of the differential geometric method and assist users to get started with the software.
\end{abstract}

\section{Introduction}
In standard robotic textbooks, orthonormal coordinate frames are used to describe robot kinematics and dynamics \cite{Craig_1986,Siciliano_2010}. 
When the Denavit-Hartenberg (DH) convention is used, predetermined rules have to be followed to position the coordinate frames and express the translational and rotational relations between them.

While this approach is popular, it has several limitations. First, multiple conventions exist to define the coordinate frames. Within these conventions, different numbers of rules have to be applied. Some conventions need special treatment, e.g., for parallel axes where the description is not unique. Second, a large number of coordinate frames has to be placed. This becomes especially unwieldy for robots with many degrees of freedom (DOF). Third, the kinematics and dynamics are expressed with one fixed set of coordinate frames on the robot bodies; if the kinematics of the robot change, e.g., for re-configurable robots, a new set of DH-parameters has to be assigned \cite{Nainer2021} and additional efforts have to be made to distinguish between revolute and prismatic joints \cite{lynch2017modern}. Fourth, the choices of task-related stationary and body-fixed frames are restricted which is disadvantageous for algorithms which describe the dynamics of multiple points on different robot bodies, e.g., for whole-body control \cite{Rocha2011}. 

In contrast, Differential Geometry can be used as a mathematical framework which lifts the coordinate-level descriptions to the more abstract space of manifolds \cite{Stramigioli1998}. Robot kinematics and dynamics can be described as actions on those manifolds \cite{Lachner_thesis_2022}. 
This mathematical abstraction leads to a formulation that requires the least number of coordinate frames to represent the robot's kinematics and dynamics.
The theoretical strengths of geometric methods have been shown in excellent textbooks \cite{Murray_1994, Selig2005, lynch2017modern} and tutorial papers \cite{Stramigioli_2001, Mueller_2018, Park2018}. Papers that compare traditional and geometric methods emphasize algorithmic and computational aspects \cite{Park_1994, Mueller_2018, Mueller_2017} but detailed discussion of  conceptual and practical differences (e.g., the brief overview in \cite{lynch2017modern}) is rare. 

Many powerful software tools exist to simulate and control robots \cite{drake, coppeliaSim, Corke_2011}. Since these tools usually offer extensive features, they present an ``overhead cost'' to learn how to use the software \cite{TodorovMuJoCo2012, Felis_2016}. This might impede first-time users, e.g., students that want to simulate a simple robot for a robotic class. 

The main contribution of this paper is a practice-oriented comparison of the traditional and  geometric approaches. The first part of the paper details the conceptual differences to derive robot kinematics and dynamics. We show that the geometric method is highly modular, flexible, and requires the least number of coordinate frames. The second part focuses on practical implementation. We introduce our software Exp[licit]\textsuperscript{TM}, a simple MATLAB robotic toolbox which leverages advantages of the geometric method. By providing Exp[licit]\footnote{https://explicit-robotics.github.io/}, we want to empower robotic researchers to experience the practical benefits of the geometric method. 

\section{Derivation of Robot Kinematics and Dynamics}\label{sec:RobotModel}
\begin{figure*}[tp]
    \centering
     \includegraphics[trim={0.0cm 9.0cm 0cm 7.0cm}, width=0.80\paperwidth,clip]{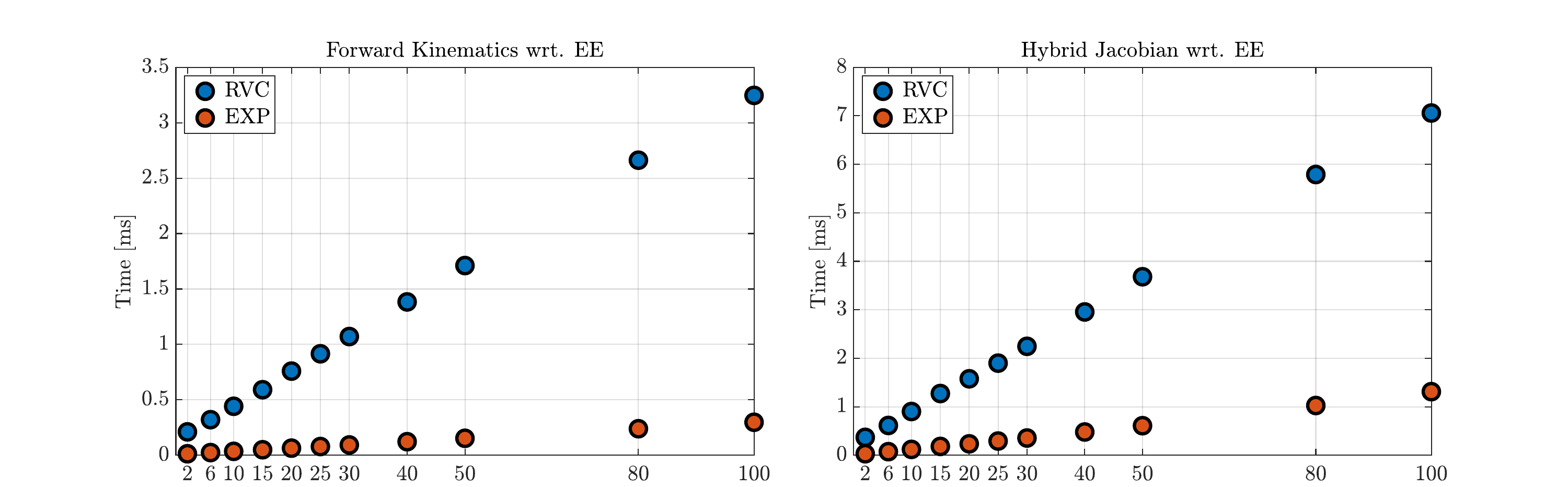}
     \includegraphics[trim={0.0cm 9.0cm 0cm 7.0cm}, width=0.80\paperwidth,clip]{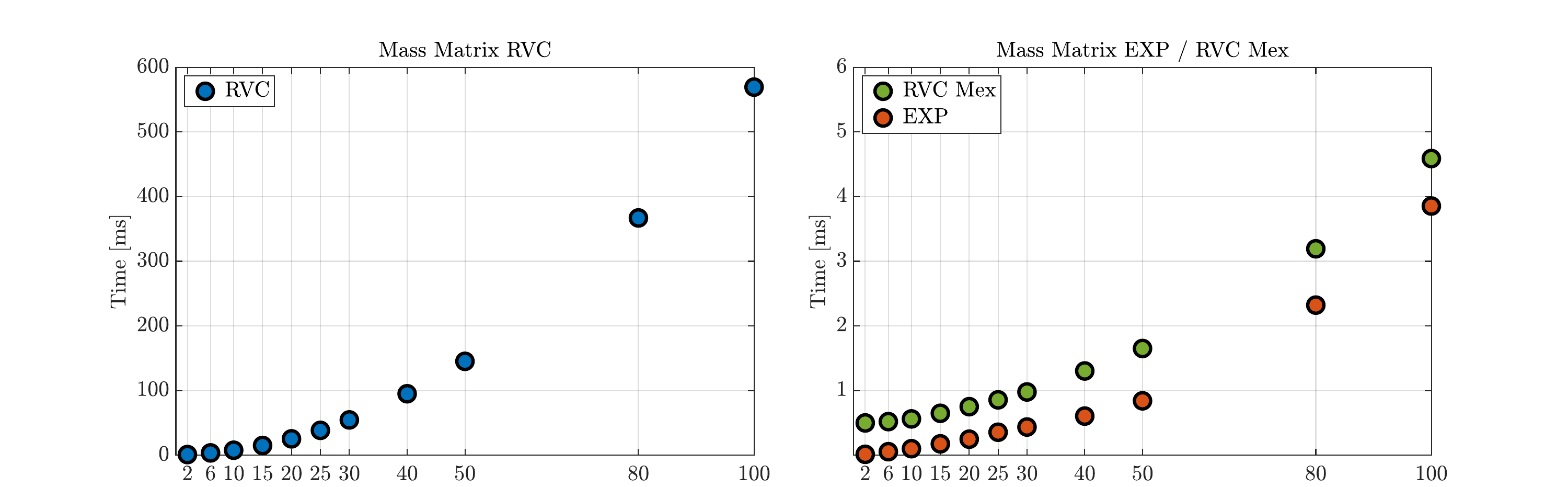}
     \includegraphics[trim={0.0cm 9.0cm 0cm 7.0cm}, width=0.80\paperwidth,clip]{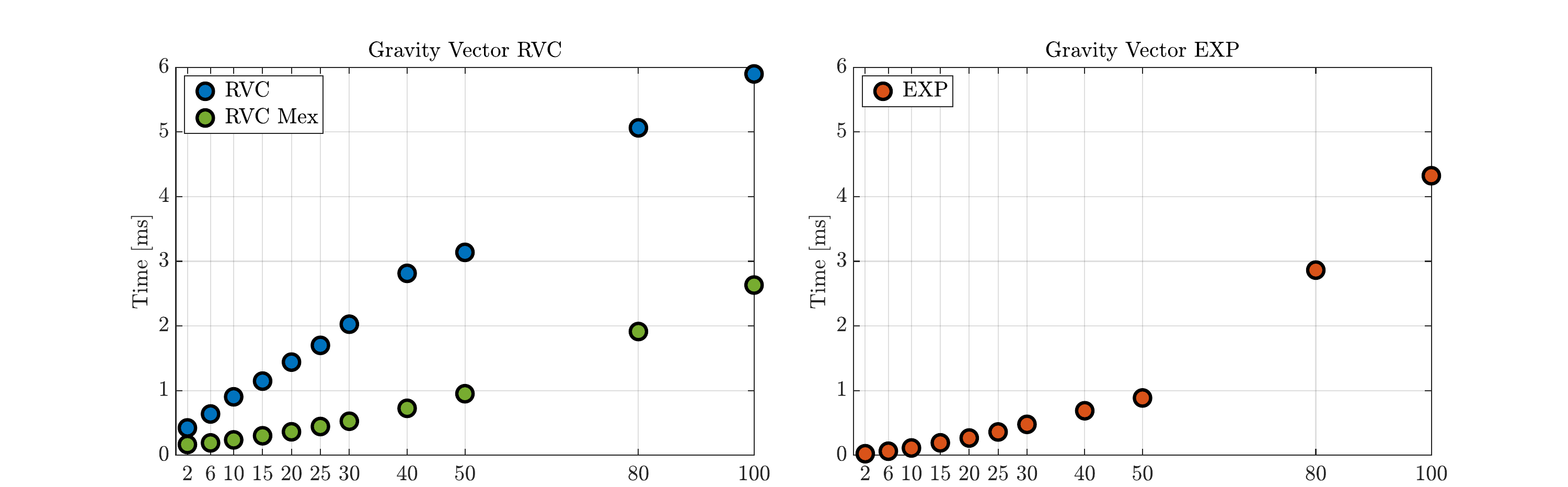}
     \includegraphics[trim={0.0cm 8.0cm 0cm 7.0cm}, width=0.80\paperwidth,clip]{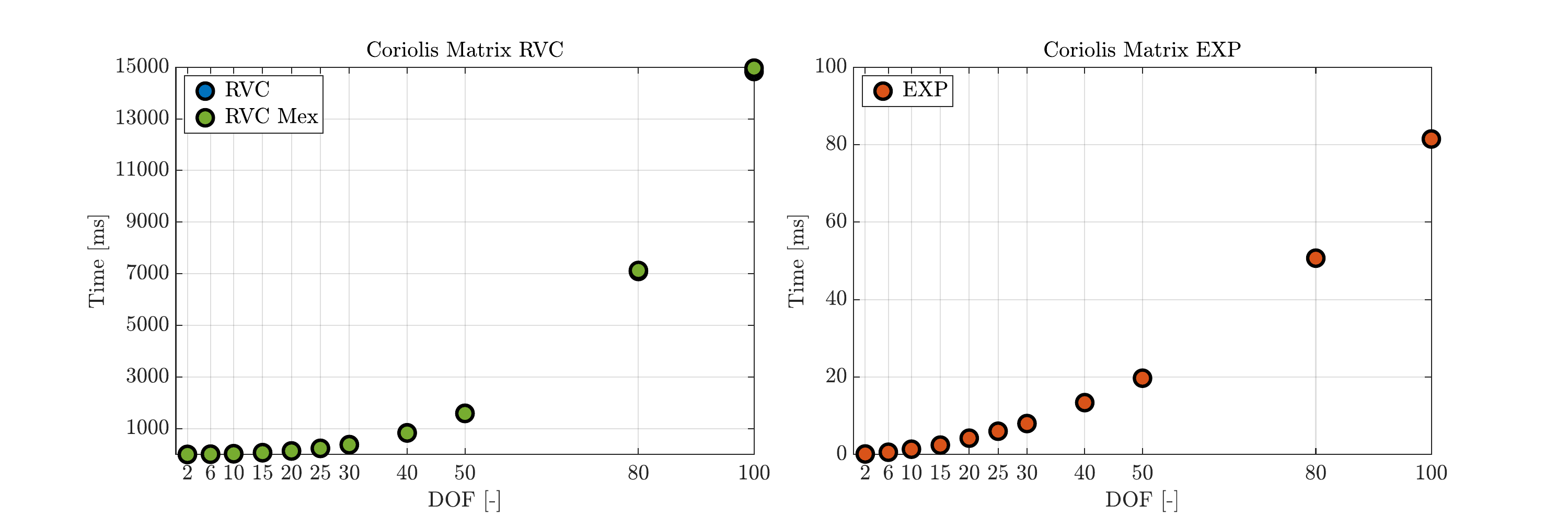}
    \caption{Computation times of RVC and Exp[licit] for five kinematic and dynamic calculations. For the Mass Matrix, Gravity, and centrifugal/Coriolis terms, the \texttt{MEX}-file option of RVC was invoked. First row: Comparison for Forward kinematics map (left) and Hybrid Jacobian (right) with respect to end-effector by using native Matlab scripts; Second row: Comparison for Mass Matrix of RVC (left) against Mass Matrix of RVC-\texttt{MEX} and Exp[licit] (right); Third row: Comparison for Gravity vector for RVC and RVC-\texttt{MEX} (left) against Gravity vector of Exp[licit] (right); Fourth row: Comparison for Coriolis Matrix of RVC (left) against Coriolis Matrix of RVC-\texttt{MEX} and Exp[licit] (right);}
    \label{fig:results_of_three_tasks}
\end{figure*}

\noindent This section shows a detailed comparison of both approaches. The theoretical derivation is focused on the Forward Kinematic Map, Jacobian matrix, and Mass Matrix of a $n$-DOF robot. A computational comparison with the RVC MATLAB toolbox\cite{Corke_2011} also includes the gravity and centrifugal/Coriolis terms (fig.~\ref{fig:results_of_three_tasks}). More details about the computational comparison are presented in sec.~\ref{subsec:compare_with_MATLAB}.

\subsection{Preliminaries}


The set of all robot configurations $\bm{q}$ constitute the manifold $\mathcal{Q}$
and the set of all homogeneous transformations $H$ constitute
the manifold $SE(3)$. To represent the robot's workspace motion, either a stationary or body-fixed coordinate frame has to be chosen.\footnote{From now on, we use ``frame(s)'' to refer to ``coordinate frame(s)''. }
We assume one stationary frame \{$S$\}, attached to the fixed base of the robot.
Moreover, we denote \{$B$\} as a body-fixed frame, which can be attached to any point of the robot. Often, \{$B$\} coincides with the tool center point (i.e., the end-effector) of the robot. In this case, we denote \{$B$\} as \{$ee$\}.

For a given joint configuration $\bm{q} \in \mathcal{Q}$, the orientation and translation of \{$ee$\} with respect to \{$S$\} can be derived via the \textit{Forward Kinematic Map}, $\mathcal{Q} \rightarrow SE(3)$ and represented by the \textit{Homogeneous Transformation Matrix} ${}^{S}{\bm{H}}_{ee}(\bm{q})= \begin{pmatrix}
{}^{S}\bm{R}_{ee} &  {}^{S} \bm{p}_{ee}\\
0 & 1
\end{pmatrix} \in SE(3)$. Here, ${}^{S}\bm{R}_{ee} \in SO(3)$ is the \textit{Rotation matrix} of \{$ee$\} with respect to \{$S$\} and ${}^{S}\bm{p}_{ee} \in \mathbb{R}^3$ is the translation from \{$S$\} to \{$ee$\}.

For a given joint motion $\dot{\bm{q}} \in \mathbb{R}^n$, the workspace motion of the robot's end-effector can be derived via the \textit{Hybrid Jacobian Matrix}\footnote{We elaborate the notion ``Hybrid'' in the next subsection. Moreover, superscript $H$ denotes ``Hybrid,'' rather than referring to a frame.} ${}^{H}\bm{J}(\bm{q}) \in \mathbb{R}^{6 \times n}$, and represented by a 6D-vector of workspace velocities, called \textit{Spatial Velocity} $^{S}{\bm{V}}_{ee} =  \begin{pmatrix}
^{S}{\bm{v}}_{ee}\\
^{S}{\bm{\omega}}
\end{pmatrix} \in \mathbb{R}^6$. Here, $^{S}{\bm{V}}_{ee}$ incorporates the linear velocity $^{S}{\bm{v}}_{ee} \in \mathbb{R}^3$ of the origin of \{$ee$\} with respect to \{$S$\} and the angular velocity $^{S}{\bm{\omega}} \in \mathbb{R}^3$ of the end-effector body, both expressed in \{$S$\}.

The total kinetic co-energy $\mathcal{L}(\bm{q},\dot{\bm{q}}) \in \mathbb{R}$ of an $n$-DOF robot is the sum of all contributions of kinetic co-energy stored by individual bodies: $\mathcal{L}(\bm{q},\dot{\bm{q}}) = \frac{1}{2} \dot{\bm{q}}^{T} \bm{M}(\bm{q})\dot{\bm{q}}$  \cite{lynch2017modern}. The matrix $\bm{M}(\bm{q}) \in \mathbb{R}^{n \times n}$ is called the \textit{Mass Matrix} of the robot.

\subsection{Traditional Method}

\subsubsection{Forward Kinematic Map via DH-convention}\label{subsubsec:Traditional_FK}
The DH-convention \cite{Denavit_1955} is widely used  to derive the Forward Kinematic Map. It is a set of rules to place body-fixed frames on the robot, and to derive the parameters that describe the kinematic relation between adjacent frames \cite{lynch2017modern}. Within the multiple DH-conventions \cite{Angeles_2006, Siciliano_Khatib__2007}, we outline the modified DH-convention which consists of four DH-parameters: link length $a$, link twist $\alpha$, link offset $d$, and joint angle $\theta$ \cite{lynch2017modern,Craig_1986, Corke_2011}.

\begin{figure}
\centering
  \includegraphics[width=1\columnwidth, clip, trim={0.0cm, 0.0cm, 0.0cm, 0.0cm}]{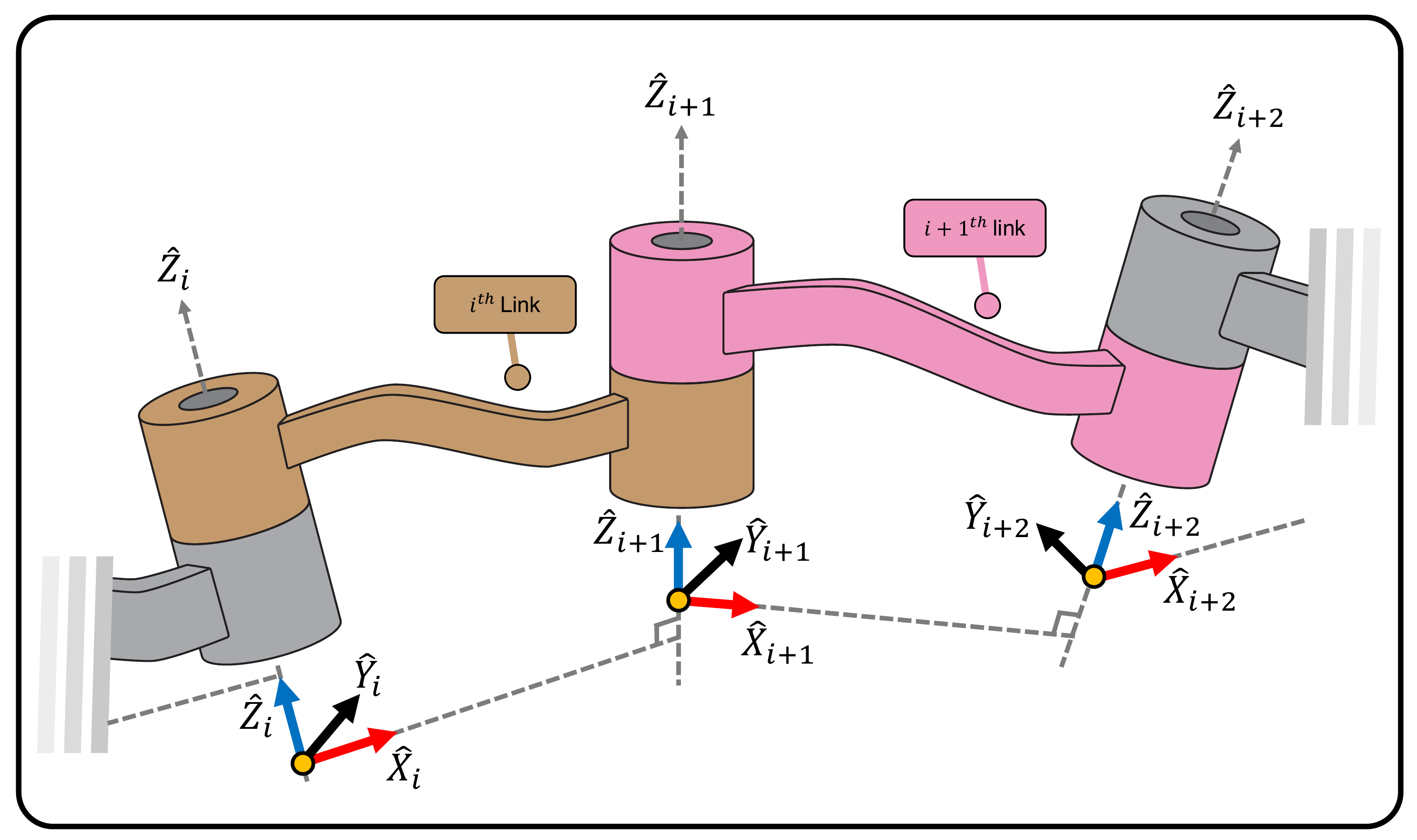}
  \caption{Frames attached to an open-chain robot, using the DH-conventions.}
  \label{fig:DH_traditional}
  \vspace{-5.0mm}
\end{figure}

To derive the DH-parameters, multiple frames have to be placed on each link using the following rules (fig.~\ref{fig:DH_traditional}):
\begin{enumerate}[label=(\roman*)]
    \item Define frames \{$1$\}, \{$2$\}, $\cdots$, \{$n$\} on each link, ordered from the base to the end-effector of the robot. Choose axis $\hat{Z}_i$ of frame $\{i\}$ to be aligned with the $i$-th joint. For a revolute (prismatic) joint, direction of $\hat{Z}_i$ is along the positive direction of rotation (translation).
    \item For $i=1,2,..., n-1$, find a line segment that is mutually perpendicular to axes $\hat{Z}_i$ and $\hat{Z}_{i+1}$. The intersection between this line and $\hat{Z}_i$ is the origin of frame $\{i\}$. Moreover, axis $\hat{X}_i$ is chosen to be aligned with this line segment, pointing from $\hat{Z}_i$ to $\hat{Z}_{i+1}$. 
    \item Attach the origin of frame \{$ee$\} to the end-effector. To simplify the derivation of the DH-parameters, the $\hat{Z}_{ee}$ axis is usually chosen to be parallel to $\hat{Z}_n$ \cite{Craig_1986}. From $\hat{Z}_n$ and $\hat{Z}_{ee}$, $\hat{X}_n$ is defined using step (ii). Finally, choose $\hat{X}_{ee}$ such that valid DH-parameters can be defined \cite{lynch2017modern}.
    \item The $\hat{Y}$ axes of frames \{$1$\}, \{$2$\}, $\cdots$, \{$n$\}, \{$ee$\} are defined using the right-hand convention. 
    \item Attach frame \{$S$\} to the robot base. Usually, it is chosen to coincide with frame \{$1$\} when joint 1 has zero displacement. 
\end{enumerate}

After assigning $n+2$ frames, \{$S$\}, $\{1\}$, $\cdots$, $\{n\}$, \{$ee$\}, the $4(n+1)$ DH-parameters can be expressed. 
With these parameters, the Homogeneous Transformation Matrix $^{i-1}{\bm{H}}_i \in SE(3)$ between frame $\{i-1\}$ and $\{i\}$ is defined for $i=1,2,...,n+1$, where $\{0\}\equiv$ \{$S$\} and $\{n+1\}\equiv$ \{$ee$\}. Finally, by concatenating these matrices, the Forward Kinematic Map, ${}^{S}\bm{H}_{ee}(\bm{q})$ can be derived:
\begin{equation}\label{eq:tradition_DH_concatenation}
    {}^{S}{\bm{H}}_{ee}(\bm{q}) = \ {}^{S}{\bm{H}}_{1}(q_1) \ {}^{1}{\bm{H}}_{2}(q_2) ... {}^{n-1}{\bm{H}}_{n}(q_n) \ {}^{n}{\bm{H}}_{ee}     
\end{equation}

\subsubsection{Jacobian Matrix by separating linear and angular velocities}\label{subsubsec:Traditional_J}
To derive the Jacobian Matrix, the traditional method separately relates joint velocities to linear and angular workspace velocities \cite{Siciliano_2010}.
We denote the linear and rotational part of the Jacobian as $\bm{J}(\bm{q})_v \in \mathbb{R}^{3 \times n}$ and $\bm{J}(\bm{q})_\omega \in \mathbb{R}^{3 \times n}$, respectively.

To derive $\bm{J}(\bm{q})_v$, the position ${}^{S}\bm{p}_{ee}$ has to be extracted from ${}^{S}\bm{H}_{ee}(\bm{q})$ (sec.~\ref{subsubsec:Traditional_FK}). Since ${}^{S}\bm{p}_{ee}$ is an analytical function of $\bm{q}$, $\bm{J}(\bm{q})_v$ collects the partial derivatives of ${}^{S}\bm{p}_{ee}$, with respect to the coordinate components of $\bm{q}$. Often,  $\bm{J}(\bm{q})_v$ is called an ``Analytical Jacobian'' \cite{Siciliano_2010}.  

The matrix $\bm{J}(\bm{q})_{\omega}$ is commonly derived  using a geometric method and  specifying the frames based on DH-convention \cite{Siciliano_2010} (sec.~\ref{subsubsec:Traditional_FK}). 
More specifically, for $i=1,2,..., n$:
\begin{itemize}
    \item If the $i$-th joint is a revolute joint with unit-rotation axis ${}^{i}\hat{\bm{\omega}}_i$ expressed in \{$i$\}, the $i$-th column of $\bm{J}(\bm{q})_{\omega}$ is ${}^{S}\bm{R}_i{}^{i}\hat{\bm{\omega}}_i = {}^{S}\hat{\bm{\omega}}_i$. 
    \item If the $i$-th joint is a prismatic joint, the $i$-th column of $\bm{J}(\bm{q})_{\omega}$ is a zero vector. 
\end{itemize}

To calculate the spatial velocity $^{S}{\bm{V}}_{ee}$, $\bm{J}(\bm{q})_{v}$ and $\bm{J}(\bm{q})_{\omega}$ can be vertically concatenated:
\begin{equation}
    {}^{S}{\bm{V}}_{ee} = {}^{H}\bm{J}(\bm{q}) \ \dot{\bm{q}}
\end{equation}
Due to the analytical derivation of $\bm{J}(\bm{q})_{v}$ and the geometrical derivation of $\bm{J}(\bm{q})_{\omega}$, we call ${}^{H}\bm{J}(\bm{q})$ the \textit{Hybrid Jacobian Matrix}. 

\subsubsection{Mass Matrix via Hybrid Jacobians}
To derive the Mass Matrix of the robot, it is necessary to attach $n$ additional frames to the center of mass (COM) of the $n$ bodies. 
These will be denoted as \{$C_1$\}, \{$C_2$\}, $\cdots$, \{$C_n$\}, ordered from the base to the end-effector of the robot. The moment of inertia of the $i$-th body with respect to \{$C_i$\} is denoted ${}^{i}\bm{\mathcal{I}}_{i}\in\mathbb{R}^{3\times 3}$. To express ${}^{i}\bm{\mathcal{I}}_{i}$ in \{$S$\}, the rotation matrix $^{S}\bm{R}_i $ is used (sec.~\ref{subsubsec:Traditional_FK}): ${}^{S}\bm{\mathcal{I}}_{i} = \ {}^{S}\bm{R}_i \ {}^{i}\bm{\mathcal{I}}_{i} \ {^{S}\bm{R}_i}^T$. 

For each body $i$, the Hybrid Jacobian Matrix ${}^{H}\bm{J}_i(\bm{q})$ is derived to describe the linear and angular velocity of \{$C_i$\} with respect to \{$S$\} (sec.~\ref{subsubsec:Traditional_J}). Note that for each matrix ${}^{H}\bm{J}_i(\bm{q})$, the columns from $i+1$ to $n$ are set to be zero since they do not contribute to the motion of body $i$ \cite{Siciliano_2010}. 

Finally, for a given mass $m_i\in\mathbb{R}$ of the $i$-th body, $\bm{M}(\bm{q}) \in \mathbb{R}^{n \times n}$ can be calculated by:
\begin{equation}\label{eq:traditional_mass_matrix}
\begin{split}
    \bm{M}(\bm{q}) = & \ m_i\sum_{i = 1}^n {{\bm{J}_i(\bm{q})_v}}^T \ {\bm{J}_i(\bm{q})_v} \\
    & + \sum_{i=1}^{n}{{\bm{J}_i(\bm{q})_\omega}}^T \ ^{S}\bm{\mathcal{I}}_{i} \ {\bm{J}_i(\bm{q})_\omega}
\end{split}
\end{equation}

\subsection{Differential geometric method}

\subsubsection{Forward Kinematic Map via the Product of Exponentials Formula}\label{subsubsec:Geometrical_FK}
For the geometric method, only two frames \{$S$\} and \{$ee$\} have to be chosen and assigned to the initial joint configuration of the robot $\bm{q}_0\in\mathcal{Q}$. The initial Homogeneous Transformation Matrix is denoted ${}^{S}{\bm{H}}_{ee}(\bm{q}_0) \equiv {}^{S}{\bm{H}}_{ee,0} \in SE(3)$. In practice it is useful to select \{$S$\} and \{$ee$\} to have equal orientation (i.e., rotation matrix equals the identity matrix) such that only the translation between \{$S$\} and \{$ee$\} has to be identified to calculate $^{S}{\bm{H}}_{ee,0}$.

In the next step, the \textit{Unit Joint Twists}\footnote{For simplicity, we will omit the term ``Unit'' in what follows.} ${}^{S}\hat{\bm{\eta}}_{i}\in \mathbb{R}^{6}$ of each joint at initial joint configuration are expressed with respect to \{$S$\}. Depending on the type of the $i$-th robot joint, ${}^{S}\hat{\bm{\eta}}_{i}\in \mathbb{R}^{6}$ is defined by: 
\begin{itemize}
    \item If the $i$-th joint is a revolute joint, the unit-axis of rotation is ${}^{S}\hat{\bm{\omega}}_i$. \textit{Any point} ${}^{S}\bm{p}_{\eta_i} \in \mathbb{R}^3$ along ${}^{S}\hat{\bm{\omega}}_i$ can be selected to define ${}^{S}\hat{\bm{\eta}}_{i}=(-[{}^{S}\hat{\bm{\omega}}_i] {}^{S}\bm{p}_{\eta_i},\ {}^{S}\hat{\bm{\omega}}_i)^T$. Here, $[{}^{S}\hat{\bm{\omega}}_i]\in so(3)$ is the skew-symmetric matrix form of ${}^{S}\hat{\bm{\omega}}_i$ \cite{lynch2017modern}. The operation $[{}^{S}\hat{\bm{\omega}}_i] {}^{S}\bm{p}_{\eta_i}$ is equal to ${}^{S}\bm{\omega} \times {}^{S}\bm{p}_{\eta_i}$.
    \item If the $i$-th joint is a prismatic joint, the unit-axis of translation is ${}^{S}\hat{\bm{v}}_i$ and therefore ${}^{S}\hat{\bm{\eta}}_{i}=({}^{S}\hat{\bm{v}}_i, \bm{0})$.
\end{itemize}
Note that the $n$ Joint Twists ${}^{S}\hat{\bm{\eta}}_{i}$ are defined with respect to a single frame \{$S$\}. For most robots, the unit-axes of rotation (or translation) can be identified by visual inspection. The positions ${}^{S}\bm{p}_{\eta_i}$ can be determined by using CAD-programs.

Finally, the \textit{Product of Exponentials Formula} \cite{Brockett_1984} can be used to derive the Forward Kinematic Map:
\begin{equation}\label{eq:geometric_POE}
\begin{split}
    ^{S}{\bm{H}}_{ee}(\bm{q}) = & \exp{([{}^{S}\hat{\bm{\eta}}_1] q_1)} \ \exp{([{}^{S}\hat{\bm{\eta}}_2] q_2)} \\
    & \cdot \cdot \cdot \exp{([{}^{S}\hat{\bm{\eta}}_{n}] q_{n})} \ ^{S}{\bm{H}}_{ee,0}
\end{split}
\end{equation}
In this equation, $[{}^{S}\hat{\bm{\eta}}_i]\in se(3)$ is a $4\times4$ matrix representation of ${}^{S}\hat{\bm{\eta}}_i$ \cite{Murray_1994}.
Given ${}^{S}\hat{\bm{\eta}}=({}^{S}\bm{v},{}^{S}\hat{\bm{\omega}})$ and $q\in\mathbb{R}$, a closed-form solution of $\exp([{}^{S}\hat{\bm{\eta}}]q)$ can be formulated \cite{lynch2017modern}:
\begin{equation}
\begin{split}
& \exp({[{}^{S}\hat{\bm{\omega}}]q}) = \mathbb{I}_3 + \sin q  [{}^{S}\hat{\bm{\omega}}] + (1-\cos q) [{}^{S}\hat{\bm{\omega}}]^2 \\
 & \bm{G}(q) = \mathbb{I}_3q + (1-\cos q)[{}^{S}\hat{\bm{\omega}}] + (q-\sin q)[{}^{S}\hat{\bm{\omega}}]^2 \\ 
& \exp{([{}^{S}\hat{\bm{\eta}}] q)} = \begin{bmatrix}
    \exp( [{}^{S}\hat{\bm{\omega}}]q ) & \bm{G}(q) {}^{S}\bm{v} \\
                        \bm{0} &  1
\end{bmatrix}\\
\end{split}
\end{equation}

\subsubsection{Jacobian Matrices via The Adjoint Map}\label{subsubsec:Geometrical_J}
 For the geometric method, two Jacobian matrices exist: the Spatial Jacobian ${}^{S}\bm{J}(\bm{q})\in\mathbb{R}^{6\times n}$ and the Body Jacobian ${}^{B}\bm{J}(\bm{q})\in\mathbb{R}^{6\times n}$ \cite{Murray_1994}. 
The Spatial (respectively Body) Jacobian relates joint velocities $\dot{\bm{q}}$ to the Spatial (respectively Body) Twist ${}^{S}\bm{\xi}$ (${}^{B}\bm{\xi}$) \cite{Murray_1994, lynch2017modern}:
\begin{equation}\label{eq:spatial_body_twists}
    {}^{S}\bm{\xi} = 
    \begin{bmatrix}
        {}^{S}\bm{v}_s \\
        {}^{S}\bm{\omega}
    \end{bmatrix} 
    =  {}^{S}\bm{J}(\bm{q})\dot{\bm{q}} 
    ~~~~~~~ 
    {}^{B}\bm{\xi} = 
    \begin{bmatrix}
        {}^{B}\bm{v}_b \\
        {}^{B}\bm{\omega}
    \end{bmatrix} 
    = {}^{B}\bm{J}(\bm{q})\dot{\bm{q}}
\end{equation}
Here, ${}^{S}\bm{\omega}$ (respectively ${}^{B}\bm{\omega}$) is the angular velocity of the body, expressed in \{$S$\} (respectively \{$B$\}); ${}^{S}\bm{v}_s$ is \textit{not} the velocity of the origin of \{$S$\}, which is zero; it is the linear velocity of a point on the robot structure, viewed as if it travels through the origin of \{$S$\} \cite{Murray_1994, lynch2017modern}; ${}^{B}\bm{v}_b$ is the velocity of the origin of \{$B$\} with respect to \{$S$\}, expressed in \{$B$\} \cite{Murray_1994, lynch2017modern}.

The columns of ${}^{S}\bm{J}(\bm{q})$ and ${}^{B}\bm{J}(\bm{q})$ are derived using the Joint Twists $\hat{\bm{\eta}}_i$ and the \textit{Adjoint Map} $\bm{Ad_H}: \mathbb{R}^{6} \rightarrow \mathbb{R}^{6}$ associated with $\bm{H} \in SE(3)$ \cite{Murray_1994, stramigioli2001modeling, lynch2017modern}. In matrix notation, $\bm{Ad_H} = \begin{pmatrix}
\bm{R} & [\bm{p}] \bm{R}\\
\bm{0} & \bm{R}
\end{pmatrix}$.  

For planar robots, $\bm{\eta}_i'$ can be identified by visual inspection. In general, 
the $i$-th column $\bm{\eta}_i'$ of ${}^{S}\bm{J}(\bm{q})$ is:
\begin{equation}\label{eq:spatial_joint_twists}
\bm{\eta}_i' = 
    \begin{cases}
        {}^{S}\hat{\bm{\eta}}_1 &\text{$i=1$} \\
        \bm{Ad}_{{}^{S}\bm{H}_{i-1}} {}^{S}\hat{\bm{\eta}}_i &\text{$i=2, ..., n$} 
    \end{cases}
\end{equation}
In this equation, ${}^{S}\bm{H}_{i-1}$ can be derived via the Product of Exponentials Formula, i.e., ${}^{S}\bm{H}_{i-1}=\exp{([{}^{S}\hat{\bm{\eta}}_1] q_1)}  \exp{([{}^{S}\hat{\bm{\eta}}_2] q_2)} \cdots \exp{([{}^{S}\hat{\bm{\eta}}_{i-1}] q_{i-1})}$.

With \{$B$\} attached to the $j$-th body, the $i$-th column $\bm{\eta}_i^{\dagger}$ of ${}^{B}\bm{J}(\bm{q})$ for $i\le j$ is:
\begin{equation}\label{eq:body_joint_twists}
\bm{\eta}_i^{\dagger} = 
     \big( \bm{Ad}_{ {}^{i}\bm{H}_{j} {}^{S}\bm{H}_{B,0} } \big)^{-1} \ {}^{S}\hat{\bm{\eta}}_i 
\end{equation}
As for eq.~\eqref{eq:spatial_joint_twists}, ${}^{i}\bm{H}_{j}$ can be derived via the Product of Exponentials Formula. Matrix ${}^{S}\bm{H}_{B,0} \in SE(3)$ is the Homogeneous Transformation of \{$B$\} with respect to \{$S$\} at initial joint configuration $\bm{q}_0$. For $j=1,2,..., n-1$, the columns of ${}^{B}\bm{J}(\bm{q})$ from $j+1$ to $n$ are zero.

\subsubsection{Mass Matrix---Mapping Generalized Inertia with Body Jacobians}
For the geometric method, the translational and rotational body contributions do not have to be separated. Instead, using the $n$ frames \{$C_1$\}, \{$C_2$\}, $...$, \{$C_n$\} (sec.~\ref{subsubsec:Traditional_J}), we define their corresponding Body Jacobian Matrices ${}^{B}\bm{J}_1(\bm{q})$, ${}^{B}\bm{J}_2(\bm{q})$, $...$, ${}^{B}\bm{J}_n(\bm{q})$ (sec.~\ref{subsubsec:Geometrical_J}). Moreover, we use $m_i$ and ${}^{i}\bm{\mathcal{I}}_{i}$ to define the \textit{Generalized Inertia matrix} $\bm{\mathcal{M}}_i = \begin{pmatrix}
       m_i \mathbb{I}_3 & \mathbf{0}\\
       \mathbf{0} & {}^{i}\bm{\mathcal{I}_{i}}
\end{pmatrix} \in \mathbb{R}^{6 \times 6}$ for each body $i$. In practice, \{$C_i$\} are aligned with the principal moments of inertia. Hence, $\bm{\mathcal{M}}_i$ can be identified by using CAD-programs.
Finally, the robot Mass Matrix can be calculated by:
\begin{equation}\label{eq:geometric_mass_matrix}
    \bm{M}(\bm{q}) = \sum_{i = 1}^n \ ^{B}\bm{J}_i(\bm{q})^T \ \bm{\mathcal{M}}_i \ ^{B}\bm{J}_i(\bm{q}).
\end{equation}

\section{Exp[licit]: Concept, Features and Use-Cases} 
This section is split into two parts. First, we highlight the conceptual and practical differences between the traditional and geometric methods. To demonstrate the practical differences, we use a Franka robot.\footnote{\url{https://www.franka.de/}} Second, we introduce Exp[licit], a MATLAB-based robot software which leverages the advantages of the geometric method. By using Exp[licit], the model parameters of the Franka robot can be derived. The modular structure of Exp[licit] will be described by using code snippets and an example application. Finally, we compare the computational efficiency of Exp[licit] with the MATLAB-based open-source robotics software ``Robotics, Vision and Control'' (RVC) which is based on the DH-convention \cite{Corke_2011}. 

\subsection{Conceptual and practical comparison between traditional and geometric methods}\label{subsec:Comparison}
\subsubsection{Forward Kinematic Map}
The DH-convention provides a minimal parameter representation (four parameters) to define the Homogeneous Transformation Matrix \cite{lynch2017modern}. This comes at a cost: a set of rules has to be carefully stipulated, which requires an extensive preparation in placing and transforming $n+2$ frames. If adjacent axes intersect or are parallel to each other, additional rules have to be considered to handle these exceptions for step (ii) in Section~\ref{subsubsec:Traditional_FK} \cite{Craig_1986}. Since rotations and translations are only allowed along/about axes $\hat{X}$ and $\hat{Z}$, the choices for frames \{$S$\} and \{$ee$\} are restricted.

In contrast, the geometric method requires only two frames: the fixed inertial frame \{$S$\} and the body-fixed frame \{$B$\}. Compared to the DH-approach, there are no restrictions on their position and orientation. The Product of Exponentials Formula provides considerable flexibility. To calculate the Joint Twists at initial configuration, any point on the twist axis can be chosen (sec.~\ref{subsubsec:Geometrical_FK}). Once the Joint Twists are defined, the Forward Kinematic Map can be derived for any point on the robot structure (sec.~\ref{subsubsec:ExampleSim}). This conceptual advantage yields a reduced computation time for the Forward Kinematic Map (sec.~\ref{subsec:compare_with_MATLAB})

\begin{figure*}[tp]
\centering
  \includegraphics[width=0.65
\textwidth, clip, trim={0.0cm, 0.0cm, 0.0cm, 0.0cm}]{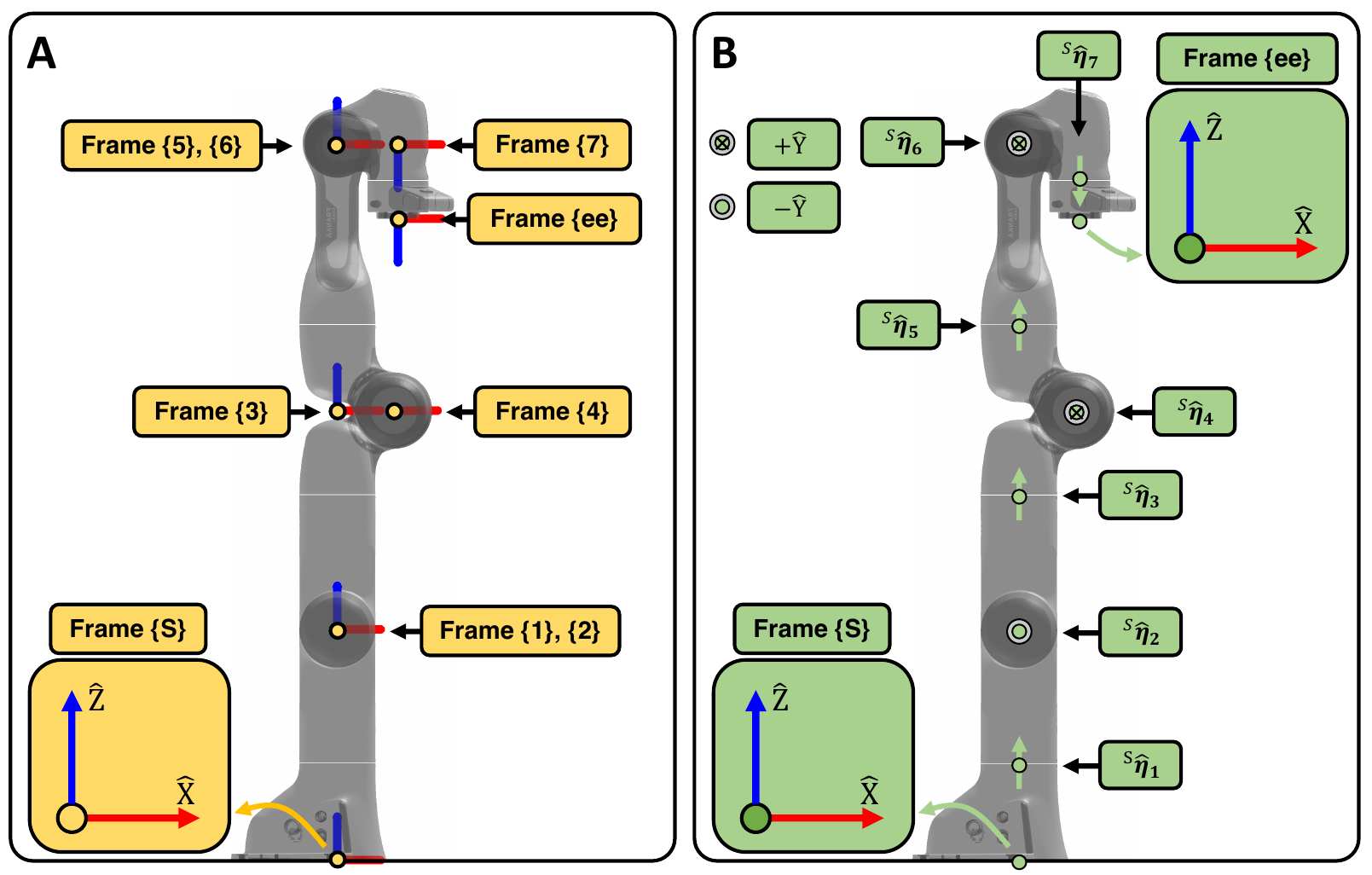}
  \caption{Franka robot at initial configuration. The DH-convention is shown in (A) and the geometric method in (B). Only two frames are required for the geometric method (B). The frames shown in (A) are derived from \cite{franka_url}.}
  \label{fig:franka_DH_vs_EXP}
  \vspace{-4.0mm}
\end{figure*}

The practical benefit of the geometric method for the Franka robot can be seen in fig.~\ref{fig:franka_DH_vs_EXP}. Compared to the DH-convention with nine frames \cite{franka_url}, only two frames are needed. For our choice of initial configuration, the calculation of ${}^{S}\bm{H}_{ee,0}$ is straightforward since only the position of the end-effector has to be calculated. For our example, ${}^{S}\bm{p}_{ee,0}=(0.088, 0, 1.033)$ and ${}^{S}\bm{R}_{ee,0}=\mathbb{I}_3$. 

The Joint Twists of the Franka robot are shown in the appendix. For a robot with revolute joints, the geometric approach needs at most four parameters (three translations parameters and one rotational parameter) like the DH-approach. For prismatic joints, the geometric approach needs only three parameters. 

\subsubsection{Jacobian Matrices}\label{subsubsec:conceptual_diff_J}
For the traditional method, the Hybrid Jacobian Matrix ${}^{H}\bm{J}(\bm{q})$ is separated into linear and angular parts. Before the linear part of ${}^{H}\bm{J}(\bm{q})$ can be derived, a choice for end-effector frame \{$ee$\} has to be made. Changing the frame at a later stage will need a recalculation of position, extracted from the Forward Kinematic Map.

The geometric approach derived two different Jacobian matrices, ${}^{S}\bm{J}(\bm{q})$ and ${}^{B}\bm{J}(\bm{q})$. The basis of the derivation are the Joint Twists at initial configuration. Hence, no separation into linear and rotational parts is needed. ${}^{S}\bm{J}(\bm{q})$ and it's output ${}^{S}\bm{\xi}$ (eq.~\eqref{eq:spatial_body_twists}) only depend on one frame \{$S$\}. By using the Adjoint Map, ${}^{S}\bm{\xi}$ can be mapped to any point on the robot structure. By choosing a point equal to the origin of \{$ee$\}, the Spatial Velocity can be derived:
\begin{equation}\label{eq:SpatialTwistToV}
            {}^{S}{\bm{V}}_{ee} = \underbrace{ \begin{pmatrix}
            \mathbb{I}_3  & - [{}^{S}\bm{p}_{ee}]\\
            \bm{0}  & \mathbb{I}_3
        \end{pmatrix} \ {}^{S}\bm{J}(\bm{q})}_{{}^{H}\bm{J}(\bm{q})} \ \dot{\bm{q}}.
\end{equation}
Here, no modification of the Forward Kinematic Map is needed, which improves the length and clarity of the code and reduces the computation time of $^{H}\bm{J}(\bm{q})$ (sec.~\ref{subsec:compare_with_MATLAB}). 

\subsubsection{Mass Matrix}
For both approaches, the frames \{$C_i$\} have to be attached to the COM of the robot at initial configuration. For the traditional method, the orientation of these coordinate frames is restricted to obtain a valid set of DH-parameters. Commonly, \{$C_i$\} is chosen to be aligned with frame \{$i$\} (fig.~\ref{fig:franka_DH_vs_EXP}A) 
and separately rotated by ${}^{S}\bm{\mathcal{I}}={}^{S}\bm{R}_{i}{}^{i}\bm{\mathcal{I}}{}^{S}\bm{R}_{i}^{T}$. 


For the geometric approach, the orientation of body frames \{$C_1$\}, \{$C_2$\}, $...$, \{$C_n$\} can be freely chosen. For each COM, the Body Jacobians are derived, again using the Adjoint Map (eqs.~\eqref{eq:spatial_joint_twists}, \eqref{eq:body_joint_twists}). 

While the traditional method divides the derivation into linear and rotational contributions, the geometric method uses the generalized inertia matrices $\bm{\mathcal{M}}_i$ (eq.~\eqref{eq:geometric_mass_matrix}) to derive the Mass Matrix. Even though $\bm{\mathcal{M}}_i$ may not be aligned with \{$S$\}, it need not be separately transformed. The transformation is incorporated in the map ${}^{B}\bm{J}_i(\bm{q}).$

\subsection{Exp[licit]---Robot modeling based on Exponential Maps}\label{subsec:using_explicit} 

The software can be installed from our Github repository: \href{https://github.com/explicit-robotics/Explicit-MATLAB}{https://github.com/explicit-robotics/Explicit-MATLAB/}. The documentation of the software can be found here: 
    \href{https://explicit-robotics.github.io/}{https://explicit-robotics.github.io/}.

\subsubsection{Software structure}
The core of the software is the \texttt{RobotPrimitives}-class, which is used as the parent class of the software. It provides the member functions \texttt{getForwardKinematics}, \texttt{getSpatialJacobian}, \texttt{getHybridJacobian}, \texttt{getBodyJacobian}, \texttt{getMassMatrix}, \texttt{getGravityVector}, and \texttt{getCoriolisMatrix} for deriving the robot parameters. 
By inheriting the \texttt{RobotPrimitives}-class, a new robot class can be defined that shares the attributes and the member functions of the parent class. Each robot class brings its kinematic and dynamic properties (e.g., axes of rotation, link lengths, masses, etc.). 

\subsubsection{Initialization}
\begin{figure*}[h]
\centering
  \includegraphics[width=1
\textwidth, clip, trim={0.0cm, 0.0cm, 0.0cm, 0.0cm}]{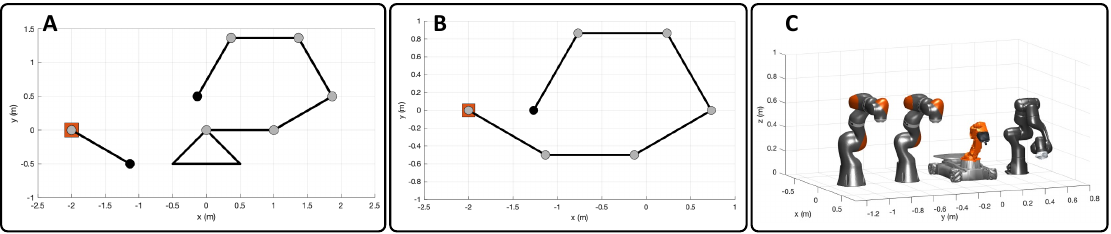}
  \caption{Exp[licit] supports various 2D and 3D-robots. (A) Two planar robots: a Cart-Pole (left) and a Snake-Robot with variable DOF (right). (B) Two robots can be combined by using the \texttt{addKinematics}-method of the \texttt{RobotPrimitives}-class. In the example (B), the two robots of (A) are combined. (C) Currently supported 3D-robots: KUKA LBR iiwa (7 and 14 kg), YouBot, and Franka.}
  \label{fig:allRobots}
  \vspace{-3.0mm}
\end{figure*}
Exp[licit] supports various 2D and 3D-robots (fig.~\ref{fig:allRobots}). In this paper, we will use a Franka robot example (\texttt{franka.m}), which is inherited from the \texttt{RobotPrimitives}-class. 
The initialization is shown below:
\begin{lstlisting}[style=Matlab-editor]
% Call Franka Robot
robot = franka( );
robot.init( );
\end{lstlisting}
The \texttt{init}-function initializes all Joint Twists and Generalized Mass Matrices for the initial configuration (fig.~\ref{fig:franka_DH_vs_EXP}). 

\subsubsection{Symbolic member functions}
All member functions also accept symbolic arguments. This feature is helpful for control methods that require an analytical formulation of the robot's equations of motion, e.g., adaptive control methods \cite{slotine1987adaptive}. An example to read out the symbolic form of the Forward Kinematics Map can be seen below:
\begin{lstlisting}[style=Matlab-editor]
% Create symbolic column vector
q_sym  = sym('q', [ robot.nq, 1 ]);                                  

% Symbolic form of Hom. Trans. Matrix
H_ee_sym = robot.getForwardKinematics( q_sym );
\end{lstlisting}

\subsubsection{Visualization and Animation}
For visualization, the robot object can be passed to a 2D or 3D-animation object:
\begin{lstlisting}[style=Matlab-editor]
% Create animation
anim = Animation('Dimension', 3, 'xLim', [-0.7,0.7], 'yLim', [-0.7,0.7], 'zLim', [0,1.4]);
anim.init( );
anim.attachRobot( robot ) 
\end{lstlisting}

The \texttt{Animation}-class heavily relies on MATLAB graphic functions (e.g., axes, patches, lighting). 
The key to our animation is to create a chain of transform objects (\texttt{hgtransforms}) instead of transforming vertices. 
The \texttt{Animation}-class has an optional input that allows the recording of videos with adjustable playback speeds.

At run-time (simulation time \texttt{t}), the robot object (in configuration \texttt{q}) and the animation can be updated: 
\begin{lstlisting}[style=Matlab-editor]
% Update kinematics
robot.updateKinematics(q);
anim.update(t);
\end{lstlisting}

\subsubsection{Modularity through Joint Twists}\label{subsec:Modularity}
The key to the modularity of Exp[licit] is the \texttt{setJointTwists( )}-function of the \texttt{RobotPrimitives}-class. So far, Exp[licit] supports revolute and prismatic joint types, indicated by the \texttt{JointTypes( )}-attribute. For each robot, the Joint Twists are derived from the joint directions (\texttt{AxisDirections}) and joint positions (\texttt{AxisOrigins}) in initial configuration. All member functions of the \texttt{RobotPrimitives}-class then re-use joints twists at runtime to map them from initial to current configuration (eq.~\eqref{eq:geometric_POE} for Forward Kinematics, eq.~\eqref{eq:spatial_joint_twists} for Spatial Jacobian, and eq.~\eqref{eq:body_joint_twists} for Body Jacobian and Mass Matrix).

\subsubsection{Example simulation}\label{subsubsec:ExampleSim}
By default, the simulation loop is set to be real-time. It is beneficial to structure the simulation script the following way: (1) calculation of all kinematic and dynamic robot parameters; (2) trajectory generation; (3) control law; (4) integration and update. For (1), the member functions of the robot object can be used. Parts (2) and (3) are generally user specific. For the integration (4), any integrator can be used, e.g., MATLAB’s pre-built \texttt{ode45.m}.

\begin{figure}[H]
  \includegraphics[width=\columnwidth, clip, trim={2.0cm, 8.5cm, 1.5cm, 8.2cm}]{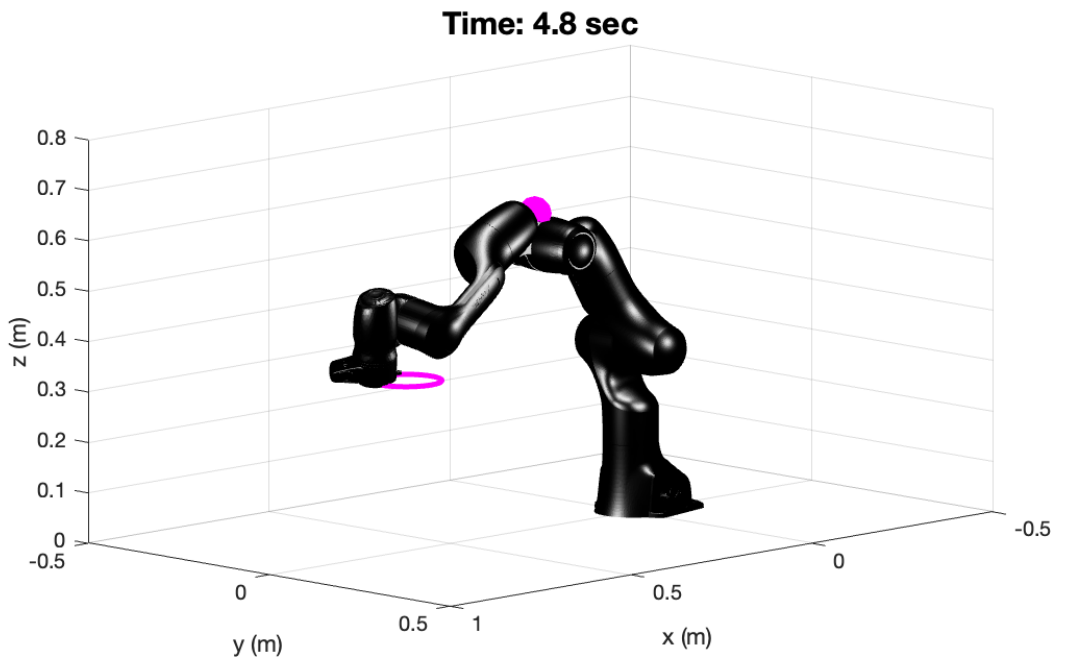}
  \caption{Simulation of a simple impedance controller, using a Franka robot.}
  \label{fig:ExampleSim}
\end{figure}

To help users with parts (2) and (3), we implemented a simple impedance controller \cite{hogan1985impedance} for a Franka robot (\texttt{main\textunderscore franka\textunderscore IC.m}) that moves the end-effector around a circular path, while keeping its elbow position (joint four) fixed (fig.~\ref{fig:ExampleSim}). 
Thanks to the modularity of the implemented geometric method, the kinematics of any point on any body can be selected by specifying the robot body (\texttt{'bodyID'}) and the corresponding position on the body (\texttt{'position'}):
\begin{lstlisting}[style=Matlab-editor]
% Get end-effector kinematics (default)
H_ee = robot.getForwardKinematics( q );
J_ee = robot.getHybridJacobian( q );

% Get kinematics of a specific point on the elbow body
H_eb = robot.getForwardKinematics( q, 'bodyID', 4, 'position', [-0.1,0,0] );
J_eb = robot.getHybridJacobian( q, 'bodyID', 4, 'position', [-0.1,0,0] );
\end{lstlisting}

\subsubsection{Comparison with MATLAB robotic toolbox}\label{subsec:compare_with_MATLAB}
We compared the computational speed of Exp[licit] with the RVC MATLAB software\cite{Corke_2011}, which uses the DH-convention. For RVC, version RTB10+MVTB4 (2017) was used.\footnote{The software can be downloaded at \url{https://petercorke.com/toolboxes/robotics-toolbox/}} 
By using native MATLAB scripts, the computation time was compared for the Forward Kinematic Map, Hybrid Jacobian, Mass Matrix, centrifugal/Coriolis terms, and Gravity vector of an $n$-DOF open-chain planar robot. The robot consisted of $n$ identical uniform-mass bars with length $l=$ 1m and mass $m=$ 1kg. While Exp[licit] calculates the gravity and centrifugal/Coriolis terms with a closed-form algorithm, RVC use recursive Newton-Euler methods (RNE). Both, Exp[licit] and RVC uses \texttt{.m}-MATLAB scripts. For the mass matrix, gravity and the centrifugal/Coriolis effects, the RVC-Method can invoke \texttt{MEX}-files to improve the computation speed. \texttt{MEX}-files are native \texttt{C} or \texttt{C++} files that are dynamically linked to the MATLAB application at runtime.

For the RVC software, the robot was constructed from the \texttt{SerialLink}-class which consists of $n$ \texttt{Revolute}-classes. For Exp[licit], the robot was constructed from the \texttt{SnakeBot}-class (fig. \ref{fig:allRobots}A). 
Robots with various DOF were constructed and tested. The test was performed with a MacBook air (M1 Chip, 16GB Memory), using MATLAB 2022a. The \texttt{timeit()} function was used to measure the computation time. 

The results of our computational comparisons are shown in Figure~\ref{fig:results_of_three_tasks}. For almost all computations, Exp[licit] was faster than the RVC software. Only for more than 70 DOF, the gravity vector of the RVC \texttt{MEX}-file option was faster than Exp[licit]. For both software, the computation of the Forward Kinematic Map and the Hybrid Jacobian showed a linear trend. The RVC software was capable of computing the Forward Kinematic Map of a 15-DOF robot within 1ms, whereas Exp[licit] required less than 0.5ms for more than 100 DOF. For the Hybrid Jacobian, the RVC software required more than 1ms for a 15-DOF robot, while Exp[licit] could accomplish the same for 80 DOF.
The computation of the Mass Matrix showed an exponential trend for both software. While Exp[licit] outperformed RCV for MATLAB scripts by a factor of 100, RVC had a much better performance using \texttt{MEX}-files. Nevertheless, it was still slower than Exp[licit]. A similar trend was seen for the gravity vector: RVC's performance was improved by invoking \texttt{MEX}-files and showed better performance for more than 70 DOF. However, for the centrifugal/Coriolis terms, Exp[licit] drastically outperformed RCV. 

These results highlight the computational advantages of a geometric approach, theoretically discussed in \cite{park1994computational}.

\section{Summary and Conclusion}
This paper summarizes and compares a traditional and a geometric method to derive the kinematic and dynamic parameters of an open-chain robot. We highlight the conceptual and practical differences between the two approaches. While the geometric method demands a more abstract perspective (i.e., mapping of Joint Twists), we showed several advantages compared to traditional methods. In summary, the advantages of the geometric method are: 1) Flexibility to express kinematic and dynamic relations without predefined rules and exceptions (sec.~\ref{subsec:Comparison}); 2) Highly modular structure, since Joint Twists can be reused throughout the calculation (sec.~\ref{subsec:Modularity}). 3) No more than two frames to describe robot kinematics and dynamics (fig.~\ref{fig:franka_DH_vs_EXP}).

We introduce Exp[licit], a MATLAB-based toolbox which implements the geometric method and leverages its advantages. Thanks to the computational advantages and highly modular structure, we believe this software can support various robotic applications. We hope to show that differential geometric methods are not limited to their conceptual strengths but can be useful for practical implementations. 

\section{Future Work}
So far, the purpose of our software is to simulate different 2D and 3D robots using MATLAB. In future, Exp[licit] will offer a \texttt{C++} and \texttt{Python} option that can be used for real-time control of robots, e.g., for torque control of cobots. At that point, it will be necessary to compare our methods with \cite{Felis_2016} which is also a library implemented in \texttt{C++}.

At the moment, Exp[licit] is limited to supporting open-chain robot structures. In the future, we are exploring the possibility of incorporating branched structures such as robotic hands, as well as closed-loop structures like delta robots.


\bibliographystyle{IEEEtran}
\bibliography{literature}


\appendices


\section*{Appendix: Parameters of the Franka robot}\label{appendix:Franka_param}
\renewcommand{\thetable}{A\arabic{table}}
\begin{table}[!ht]\label{ap:franka_DH_vs_EXP}
\centering
\includegraphics[width=0.9\columnwidth, clip, trim={0.0cm, 4.3cm, 0.0cm, 2.5cm}]{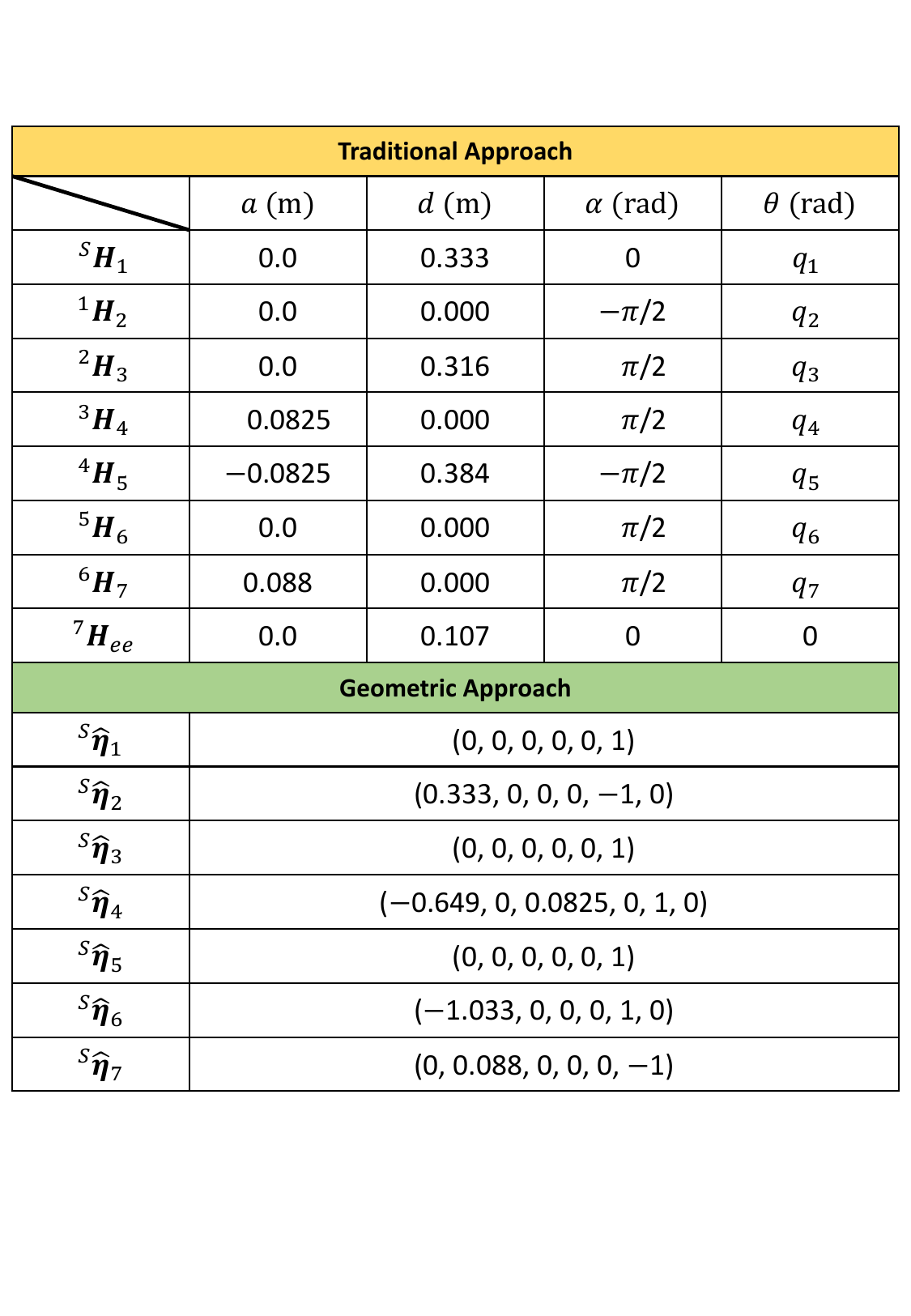}
\end{table}

\end{document}